\documentclass{article}
\usepackage{spconf,amsmath,graphicx}
\usepackage{multirow}
\usepackage{url}

\title{Improving Depression estimation from facial videos with face alignment, training optimization and scheduling}
\name{Manuel Lage Cañellas$^{\star }$, Constantino Alvarez Casado$^{\star }$, Le Nguyen$^{\star }$, Miguel Bordallo López$^{\star\dagger}$}

\address{$^{\star}$ Center for Machine Vision and Signal Analysis (CMVS), University of Oulu, Finland\\
        $^{\dagger}$ VTT Technical Research Centre of Finland}

\begin{document}
%
\maketitle

%
\begin{abstract}
Deep learning models have shown promising results in recognizing depressive states using video-based facial expressions. While successful models typically leverage using 3D-CNNs or video distillation techniques, the different use of pretraining, data augmentation, preprocessing, and optimization techniques across experiments makes it difficult to make fair architectural comparisons. We propose instead to enhance two simple models based on ResNet-50 that use only static spatial information by using two specific face alignment methods and improved data augmentation, optimization, and scheduling techniques. Our extensive experiments on benchmark datasets obtain similar results to sophisticated spatio-temporal models for single streams, while the score-level fusion of two different streams outperforms state-of-the-art methods. Our findings suggest that specific modifications in the preprocessing and training process result in noticeable differences in the performance of the models and could hide the actual originally attributed to the use of different neural network architectures.
\end{abstract}

%

\vspace{-2mm}
\section{Introduction}
\label{sec:intro}
\vspace{-2mm}
Depression is a common mental health disorder that negatively affects an individual's well-being \cite{beck1996bdi}. Long-term medical depression can lead to severe complications, both at psychological and physiological levels. Several studies suggest depression as a trigger of other diseases such as cardiovascular disease, osteoporosis, aging, pathological cognitive changes, Alzheimer's disease, and other dementias, and even an increase in the risk of earlier mortality \cite{Verhoeven2013DepressionAging}.

Systems that automatically recognize depression are desirable because of their potential objectivity, speed, and reliability to avoid such an impact on a patient's health and well-being. In the last decade, many approaches based on classical statistical machine learning algorithms have been proposed to recognize signs of depression from facial videos, speech, and text data \cite{SpeechAnalysisMDD2022Wu} to help decision-making by physicians.

While the most novel architectures have shown noticeable improvements in the accuracy of depression recognition models, most previous work does not discuss or experiment with substantial components of the machine learning pipeline, such as preprocessing or optimization. Based on these shortcomings, in this article, we propose creating deep learning models for automatic depression screening using only static textural features extracted from facial video frames. In this context, we suggest a set of changes that improve the results using this kind of architecture. Our main contribution can be summarized as follows:


\begin{itemize}
\setlength\itemsep{0pt}
\setlength\parskip{0pt}
\item We introduce a set of 2D-CNN models based on the ResNet-50 architecture \cite{resnet2016} trained using only static textural information from video frames by applying two different face alignment techniques, and evaluate their impact in the final results.

\item We explore novel training optimization and scheduling schemes to further improve the results from previous similar approaches that are based on spatial information.

\item We propose the use of a fusion score approach to regress the depression levels using different textural-based models that show to be complementary depending on the face alignment.

\item We train and validate the models on the AVEC2013 \cite{Avec2013Valstar2013} and AVEC2014 databases \cite{Avec2014Valstar2014}, indicating that this approach can obtain comparable results to sophisticated spatio-temporal models, while the score-level fusion of both streams models outperforms state-of-the-art methods in the literature.

\item Finally, we find that using only textural-based models suggests that small changes in the preprocessing and training process could result in noticeable differences in the performance of the models, which could be hiding the real the contributions attributed to the differences in the neural network architectures.

\end{itemize}

\section{Related Work}
In the last years, computer vision has been proposed as a powerful tool to diagnose clinical depression, as it shows promising performance in recognizing and analyzing facial expressions, a trait that is known to be deeply connected with depression. Several studies suggest that depression affects facial expressions in the following way: depressed people generally show a decrease in the intensity of positive emotions, a reduction of the number of smiles, an increase in the intensity of negative emotions, a reduction of the number of eye contacts, a reduction of the number of blinks, a reduction of the number of head turns and an increase in the number of head nods \cite{ClinicalDepressionFAUdifferences2018}. Many studies in the computer vision community have proposed automatic depression detection (ADD) methods based on facial expression analysis (FEA) in both static and video scenarios \cite{ReviewADD2021}. In particular, to recognize depressed expressions in static images, methods based on deep learning models have been proposed to extract embedding vectors from faces and classify images into depression or control classes \cite{ResNet50Melo2019}. Other studies propose a similar approach with static images but focus on facial multi-regions instead of the entire face \cite{DCNNZhou2018}. More recently, most studies have focused on exploiting the spatio-temporal information by leveraging the facial interframe information from videos. Some studies proposed 3D-CNN architectures to extract spatio-temporal features from short video clips, while others proposed temporal pooling techniques to capture and encode the dynamic information of video clips into an image map and train 2D-CNNs \cite{Melo2020TwoStream}.

\section{Proposed methodology}
In this paper, we propose using preprocessed RGB images to learn discriminative representations to determine the level of depression of a person. In contrast to most of the works in the literature, our work focuses on techniques for preprocessing and data augmentation of the input data, and keeps the same backbone network, a pretrained ResNet-50 architecture.

%
\subsection{Preprocessing and face alignment}
\begin{figure}[t]
  \begin{center}
    \includegraphics*[width=1\columnwidth]{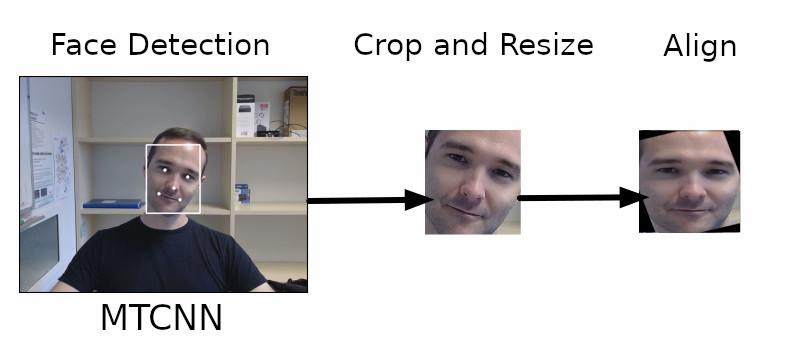}
  \end{center}
  \vspace{-5mm}
  \caption{Incorrect alignment process. Cropping the facial region before the rotation deletes textural information within the facial boundaries, depicted as black triangles.}
  \label{fig:bad_alignment}
\end{figure}

\begin{figure}[t]
  \begin{center}
    \includegraphics*[width=1\columnwidth]{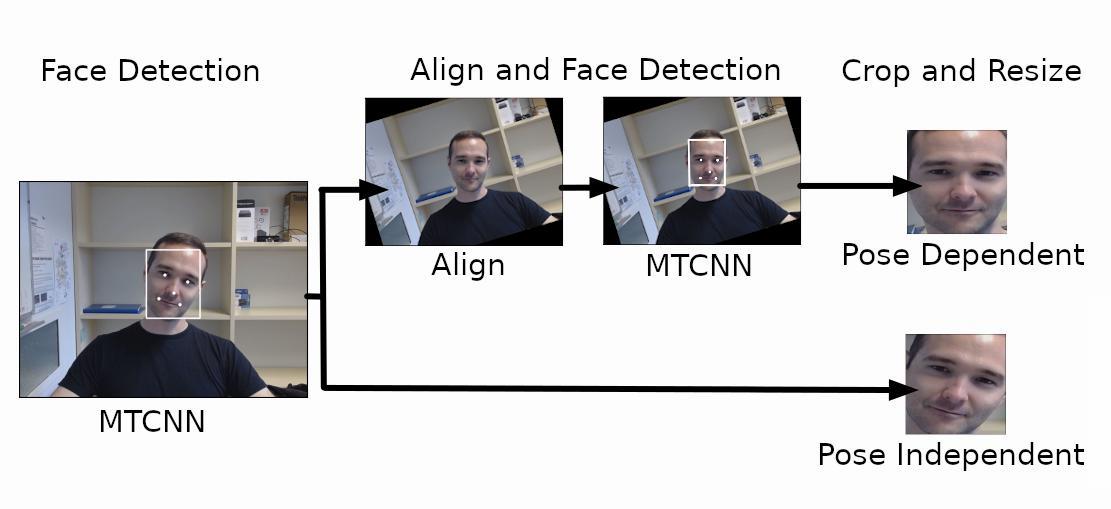}
  \end{center}
  \vspace{-5mm}
  \caption{Two different face alignment processes. Pose dependent alignment detects the face and landmarks, rotates the image aligning the eyes horizontally and crops the face boundaries with a new face detection. Pose independent alignment detects the face and landmarks and crops the face boundaries directly.  }
  \label{fig:good_alignment}
\end{figure}

The first step of our method consists in segmenting and aligning the facial regions from every video frame. We use a state-of-the-art face detector, a Multi-task Cascade Convolutional Network (MTCNN) \cite{ZhangMtcnn2016} based in deep learning, which provides also face alignment by detecting five fiducial landmarks in the eyes, nose and mouth. As seen in the related literature, \cite{Melo2020TwoStream,Melo2021mdn}, typical face preprocessing is based on cropping, aligning and re-scaling the images. However, the order of cropping and aligning faces has a direct impact on the resultant image. Cropping the facial region before the rotation and alignment deletes textural regions at the border. This can be depicted as black triangles, as shown in Figure \ref{fig:bad_alignment} discarded facial alignment. 
As our model is only based in texture information, we apply instead a preprocessing scheme that conserves all textural information within the facial boundaries. In this context, we align and rescale the images in two different manners, pose independent and pose dependent, as shown in \ref{fig:good_alignment}.

\vspace{-3mm}

\vspace{-3mm}

Pose dependent face alignment rotates the face based on the position of the eyes, which are horizontally aligned, subsequently applying a new face detection from where the aligned face is cropped and scaled, providing an input that varies the texture with the facial expression. Pose independent face alignment simply crops the face based only on the facial boundaries provided by MTCNN and then scales the image, providing an input that varies the texture with different head poses. These two types of alignments could give complementary information of the texture of the face.

\subsection{Architecture}
\begin{figure}[t]
  \begin{center}
    \includegraphics*[width=1.0\columnwidth]{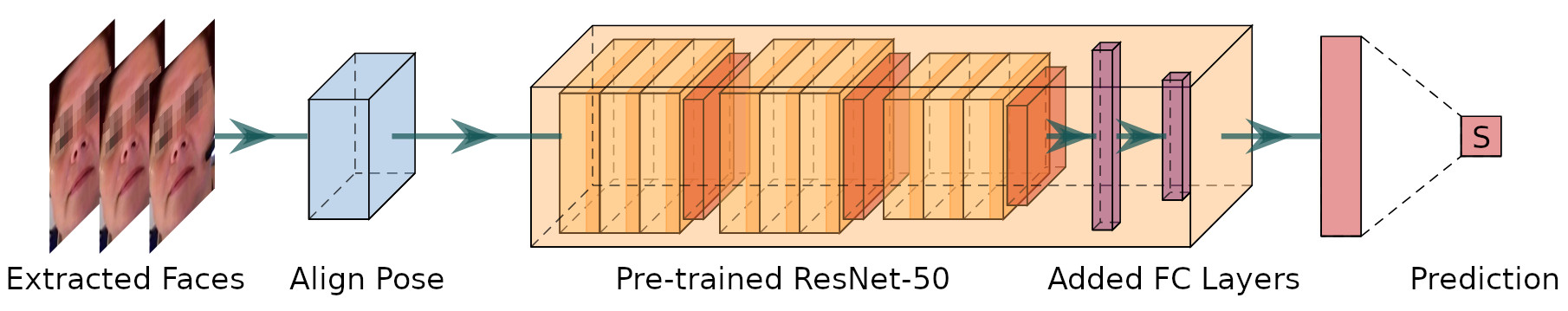}
  \end{center}
  \vspace{-5mm}
  \caption{Proposed Architecture of a ResNet-50 architecture followed by two additional fully connected layers of 512 and 128 neurons. }
  \label{fig:architecture}
\end{figure}

We use a ResNet-50 architecture \cite{resnet2016} as the backbone, followed by two additional fully connected (FC) layers of 512 neurons and a regression layer of 128 neurons to estimate the depression level as shown in Figure \ref{fig:architecture}. To take advantage of prelearnt features and transfer learning, we initialize the network with pretrained weights based on the InsightFace framework \cite{guo2021sample}.
In order to favor that new low-level textural features can be learnt from our specific data, we keep unfrozen all the layers of the architecture. The prediction \textit{S} of each video is then calculated as the average of the predictions of all individual images.

\subsection{Data Augmentation}

Since our proposed approach is based on the complementarity of information of both pose dependent and pose independent faces, we do not add any rotation of the faces for data augmentation. We propose instead random horizontal flips for each face, and changes in brightness, contrast and saturation. In contrast to other works in the literature, we do not include vertical flips nor add extra images to the dataset. We add these data augmentation techniques to our previous pose dependent and pose independent streams. 

\subsection{Training}
On the AVEC2014 database, we process every frame of each video for a total of 304929 facial frames. As AVEC2013 database is nine times bigger than AVEC2014, in order to keep a similar amount of data, and similarly to previous works, we decide to process 1 frame every 9. The cost of this regression model is calculated as the L1 norm, the mean average error function (MAE) that aims at minimizing the error for the predictions of each individual frame. In contrast with other previous publications, we include a state-of-the-art optimizer, RAdam \cite{liuRadam2019}, enhanced with an implementation of Lookahead optimization \cite{LookaheadZhang2019}. We use a ReduceLROnPlateau optimizer that reduces our initial learning rate of $3.10^{-4}$ \cite{BermantBioCPPNet2021} when learning has stopped improving.


\section{Experimental Analysis and Results}

\subsection{Datasets}
The performance of our proposed method has been evaluated on two publicly available databases: Audio/Visual Emotion Challenge AVEC2013 \cite{Avec2013Valstar2013} and AVEC2014 \cite{Avec2014Valstar2014} depression sub-challenge datasets. Both of them are derived from a subset of the audio-visual depressive language corpus (AViD-Corpus).  
AVEC2013 is constituted of one task while AVEC2014 is distributed into two different subsets: Northwind and Freeform, where the same subject performs two tasks. 
Each task is segmented into three partitions: training, development, and test, all of them with 50 videos. Every video contains a record of a subject speaking in front of a camera. The videos are labeled with a Beck-Depression Inventory BD-II \cite{beck1996bdi} score indicating a depression level that ranges between 0 and 63.
According to the BD-II score, the severity of depression can be classified into four levels: minimal (0-13), mild (14-19), moderate (20-28), and severe (29-63).

\subsection{Experimental setup}
The proposed methodologies are evaluated using only the static texture features extracted from both benchmark dataset videos.
The results across different models are compared against other state-of-the-art models based on visual information, including static and spatio-temporal models. We provide results for both for individual streams and for their complementary score-level fusion. 
The experiments are performed using a computer that integrates an AMD Ryzen 7 5800 8-Core processor and an NVIDIA GeForce RTX 3060 running on Linux. We used Python 3.8 as the programming language with PyTorch 1.11.0 framework.

\subsection{Protocol and Performance metrics}
To evaluate the performance of these models and make a fair comparison with the state-of-the-art methods, we provide the two most common metrics in the automatic depression assessment literature, Mean Absolute Error (MAE) and Root Mean Squared Error (RMSE). The overall predicted depression score for each input video is obtained by averaging the estimation scores for all its frames.

\subsection{Experimental results}

The performance and validity of the proposed modality is evaluated through a series of experiments in the benchmark databases. Table \ref{tab:res_2013} shows the results of both pose dependent and pose independent streams and the fusion of them, for AVEC2013.
In table \ref{tab:res_2014} we provide the results for AVEC2014, with two different ways for measuring the performance: separated and joint tasks. 
Separated task evaluation considers each video as an independent task. The MAE is calculated over the mean of 100 predictions, one per each video.
Joint task evaluation fuses each Northwind and Freeform video done by the same subject by averaging the predictions of both tasks. The MAE is then obtained as the mean over the 50 predictions corresponding to each subject.
For the two tables, it can be seen that the average error, in terms of MAE, the error is smaller when streams are fused. These results show that the information contained in pose dependent and pose independent streams might be complementary. Hence, the exploration of different preprocessed spatial streams is an effective way to utilize the static information.

\renewcommand{\arraystretch}{1.2}
\begingroup
\setlength{\tabcolsep}{5pt}
\begin{table}[h]
\centering

\begin{tabular}{|l|l|l|}
\hline 
Streams & MAE & RMSE\\
\hline
Pose independent & 6.16 & 8.99 \\
Pose dependent &  6.02 & \textbf{8.23} \\
\hline
Fused & \textbf{5.82} & 8.41 \\
\hline 
\end{tabular} 
\caption{Experimental results for AVEC2013.}
\label{tab:res_2013}
\end{table}
\endgroup

\renewcommand{\arraystretch}{1.2}
\begingroup
\setlength{\tabcolsep}{5pt}
\begin{table}[h]
\centering
\begin{tabular}{|l|l|l|l|l|}
\cline{2-5}
 \multicolumn{1}{c|}{} & \multicolumn{2}{c|}{Separated tasks} & \multicolumn{2}{c|}{Joint tasks} \\
\hline
Streams & MAE & RMSE & MAE & RMSE\\
\hline
Pose Independent & 6.00 & 8.08 & 5.94 & 7.88 \\  
Pose Dependent   & 5.61 & 7.52 & 5.59 & 7.34 \\ 
\hline
Fused & \textbf{5.53} & 7.42 & \textbf{5.50} & 7.29 \\ 
\hline 
\end{tabular} \\
\caption{Experimental results for AVEC2014. Comparison for Separated and Joint tasks.}
\label{tab:res_2014}
\end{table}
\endgroup

\section{Error distribution}
\begin{figure}[h]
  \begin{center}
    \includegraphics*[width=1.05\columnwidth]{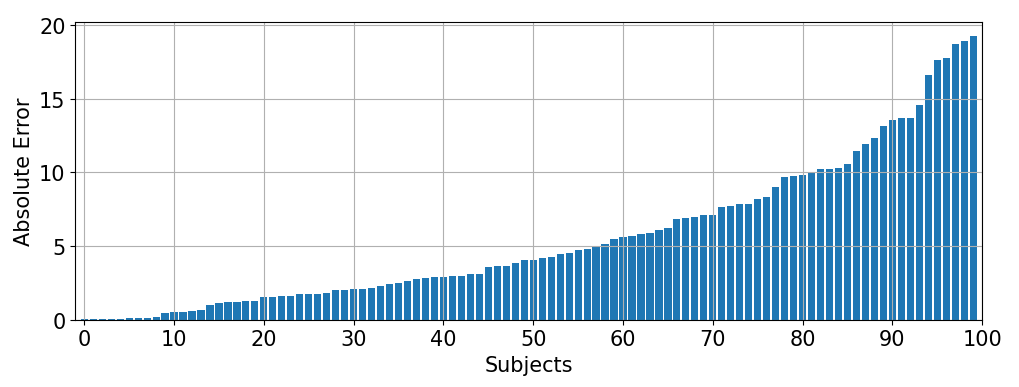}
  \end{center}
    \vspace{-5mm}
  \caption{Absolute error distribution for each video in AVEC2014}
  \vspace{-5mm}
  \label{fig:ErrorDistribution}
\end{figure}

To further analyze the performance of our best model we show the error distribution in AVEC2014 benchmark, and show it in Figure \ref{fig:ErrorDistribution}. The figure shows the absolute error for each of the 100 test videos ordered from the smallest to the largest. We can observe that the error distributions of our approach shows a relatively small overall error. Our textural model shows a higher portion of videos with a very small error (that would not result in a misclassification diagnosis). However, a small portion of the videos, still show significantly high errors. 

\subsection{Comparison with State-of-the-Art}
Tables \ref{tab:sota_2013} and \ref{tab:sota_2014}, show the evaluation of the performance of the proposed approach using different alignment and the improved pretraining, optimization and scheduling techniques applied to facial videos for both AVEC2013 and AVEC2014. We make the comparison with the result that we obtained by fusing the streams. The comparison shows how using only textural information and a well-known deep learning architecture, it is possible to obtain state-of-the-art results comparable with the most sophisticated novel architectures that exploit spatio-temporal information.

\renewcommand{\arraystretch}{1.2}
\begingroup
\setlength{\tabcolsep}{10pt}
\begin{table}[h]
\centering
\begin{tabular}{|l|l|l|}
\hline 
Methods & MAE & RMSE \\
\hline
ResNet-50+Pool (Zhou \emph{et al.} \cite{ResNet50Zhou2019})       &6.37 &8.43 \\
MTB-DFE (Xu \emph{et al.} \cite{XuTwoStages2021})                   &6.31 &8.20 \\
Four DCNN (Zhou \emph{et al.} \cite{DCNNZhou2018})                  &6.20 &8.28 \\
MDN-100 (Melo \emph{et al.} \cite{Melo2021mdn})                     &6.14 &7.62 \\
MTB-DFE + SEG (Xu \emph{et al.} \cite{XuTwoStages2021})             &6.05 &7.92 \\
2xResNet-50 (Melo \emph{et al.} \cite{Melo2020TwoStream}) &5.96 &7.97 \\
MTB-DFE + SPG (Xu \emph{et al.} \cite{XuTwoStages2021})             &5.95 &7.57 \\

\hline 
Fusion of texture streams (ours) & \textbf{5.82} & 8.41 \\
\hline 

\end{tabular}
\caption{Comparison of methods for depression detection on AVEC2013 dataset}
\label{tab:sota_2013}
\end{table}
\endgroup

\begingroup
\setlength{\tabcolsep}{10pt}
\begin{table}[h]
\centering
\begin{tabular}{|l|l|l|}
\hline 
Methods & MAE & RMSE \\
\hline
ResNet-50+Pool (Zhou \emph{et al.} \cite{ResNet50Zhou2019})       &6.37 &8.43 \\
MTB-DFE (Xu \emph{et al.} \cite{XuTwoStages2021})                   &6.30 &7.83 \\
MTB-DFE + SPG (Xu \emph{et al.} \cite{XuTwoStages2021})             &6.24 &7.65 \\
Four DCNN (Zhou \emph{et al.} \cite{DCNNZhou2018})                  &6.21 &8.39 \\
2xResNet-50 (Melo \emph{et al.} \cite{Melo2020TwoStream})           &6.20 &7.94 \\
MDN-152 (Melo \emph{et al.} \cite{Melo2021mdn})                     &6.06 &7.65 \\
MTB-DFE + SPG (Xu \emph{et al.} \cite{XuTwoStages2021})             &5.86 &7.18 \\

\hline 
Fusion of texture streams (ours) & \textbf{5.50} & 7.29 \\ 
\hline 
\end{tabular}
\caption{Comparison of methods for depression detection on AVEC2014 dataset}
\label{tab:sota_2014}
\end{table}
\endgroup

Although our results are based only on static information, they obtain the best results in terms of MAE and very competitive results in terms of RMSE. However, since we use a very simple architecture, it is possible to argue that similar preprocessing and optimization techniques applied to more sophisticated spatio-temporal models would result in an even better performance. 

\section{Conclusion}

This paper introduced a simple depression detection approach that leverages two complementary face alignment techniques to derive a set of deep models based on textural static information. In addition, the training process of the models is improved by data augmentation and novel optimization and scheduling schemes. Extensive experiments on the AVEC2013 and AVEC2014 benchmark datasets show that individual models trained this way have comparable results to sophisticated spatio-temporal models, while the score-level fusion of several streams outperforms more sophisticated state-of-the-art methods based on visual information. We argue that, at least for AVEC2013 and AVEC2014, the impact of using novel architectures in depression estimation from videos, can not be clearly distinguished from the contributions due to other processing components, and should be investigated in a more systematic manner.

\bibliographystyle{IEEEbib}
\bibliography{references}

\end{document}